\documentclass[11pt]{article}
\usepackage{acl2016}
\usepackage{times}
\usepackage{latexsym}
\usepackage{comment}
\usepackage{graphicx}
\usepackage{url}

\aclfinalcopy 

\newcommand{\GG}[1]{}

\title{Semantic Representations of Word Senses and Concepts}

\author{Jos\'e Camacho-Collados, Ignacio Iacobacci, Roberto Navigli \and Mohammad Taher Pilehvar$^\dagger$  \\
	    Department of Computer Science, Sapienza University of Rome\\
	      	 {\tt \{collados,iacobacci,navigli\}@di.uniroma1.it}\\
	$^\dagger$Language Technology Lab, DTAL, University of Cambridge\\
        {\tt     mp792@cam.ac.uk}}

\date{}

\begin{document}

\maketitle

\section{Introduction}

 Representing the semantics of linguistic items in a machine-interpretable form has been a major goal of Natural Language Processing since its earliest days. Among the range of different linguistic items, words have attracted the most research attention. However, word representations have an important limitation: they conflate different meanings of a word into a single vector. Representations of word senses have the potential to overcome this inherent limitation. Indeed, the representation of individual word senses and concepts has recently gained in popularity with several experimental results showing that a considerable performance improvement can be achieved across different NLP applications upon moving from word level to the deeper sense and concept levels. Another interesting point regarding the representation of concepts and word senses is that these models can be seamlessly applied to other linguistic items, such as words, phrases and sentences.

This tutorial\footnote{Slides available at \url{http://goo.gl/az7tBD}} will first provide a brief overview of the recent literature concerning word representation (both count and neural network based). It will then describe the advantages of moving from the word level to the deeper level of word senses and concepts, providing an extensive review of state-of-the-art systems. Approaches covered will not only include those which draw upon knowledge resources such as WordNet, Wikipedia, BabelNet or FreeBase as reference, but also the so-called multi-prototype approaches which learn sense distinctions by using different clustering techniques. Our tutorial will discuss the advantages and potential limitations of all approaches, showing their most successful applications to date. We will conclude by presenting current open problems and lines of future work.

\section{Outline}

\subsection{Semantic Representation: Foundations}

This session provides the necessary background for semantic representation. We will briefly cover the traditional vector space model \cite{TurneyPantel:10} followed by the more recent approaches based on neural networks \cite{Mikolovetal:2013}. We then provide reasons for the need to produce semantic representations for the deeper word sense level, focusing on the main limitation of the word-based approaches which is their inherent ambiguity. Finally, we show how sense-based representations are bound to overcome these limitations, hence providing improvements across several tasks.

\subsection{Knowledge-based sense representations}
    
We start this session by briefly introducing some of the most popular lexical knowledge resources that have been used by different sense representation techniques. We put emphasis on WordNet \cite{Milleretal:90}, the de facto standard sense inventory in the community, and Wikipedia, the largest collaboratively-constructed resource of the type, both of which have been extensively used by many researchers in the area. We discuss the advantages each of these resources provides and show how they are usually viewed as semantic networks and exploited for representation purposes.

Then, we provide a deep review of different techniques that learn representations for individual concepts in a target sense inventory. We cover all the existing approaches that model concepts in WordNet \cite{PilehvarNavigli:2015aij}, articles in Wikipedia \cite{HassanMihalcea:2011}, or concepts in larger sense inventories such as BabelNet \cite{iacobacci:2015,camachocollados:2016aij} or FreeBase  \cite{bordes2011learning,bordes2013translating}. We will also cover some approaches that make use of additional external corpora (or word representations learned on the basis of statistical clues) besides the target knowledge resource \cite{chenunified:2014,chen2015improving,johansson2015embedding,jauhar2015ontologically,RotheSchutze:2015,PilehvarCollier:2016}. We discuss the advantages of these knowledge-based representations and focus on how neural network-based learning has played a role in this area in the past few years.

\subsection{Unsupervised sense representations}

In this session we cover the so-called multi-prototype techniques that learn multiple representations per word, each corresponding to a specific meaning of the word. We will illustrate how these approaches leverage clustering algorithms for dividing the contexts of a word into multiple contexts for its different meanings \cite{ReisingerMooney:2010,Huangetal:2012,Neelakantanetal:2014,tian2014probabilistic,wu2015sense,li-jurafsky:2015:EMNLP,liu2015topical,vu2016k,vsuster2016bilingual}.

\subsection{Advantages and limitations}

This session reviews some of the advantages and limitations of the knowledge-based and unsupervised techniques, describing the applications for which they are suitable and mentioning some issues such as the knowledge acquisition bottleneck.

\subsection{Applications}

This session focuses on different applications of sense representations. We briefly mention some of the main applications and tasks to which sense representations can be applied. Sense representations may be used in virtually every task in which word representations have been traditionally applied. Examples of such tasks include automatic thesaurus generation \cite{Crouch:1988,CurranMoens:2002}, information extraction \cite{Laenderetal:2002}, semantic role labelling \cite{Erk:07,Pennacchiottietal:2008}, and word similarity \cite{Deerwesteretal:1990,Turney:2003,Radinsky:2011,mikolov2013distributed} and clustering \cite{PantelLin:2002}. We will provide comparisons between word and sense representations performance, discussing the advantages and limitations of each approach. Moreover, we will show how sense representations can also be applied to a wide variety of additional tasks such as entity linking and word sense disambiguation \cite{navigli:09,chenunified:2014,camachocolladosetal:2015b,RotheSchutze:2015,camachocollados:lrec2016}, sense clustering \cite{Snowetal:2007,camachocolladosetal:2015}, alignment of lexical resources \cite{niemann2011people,PilehvarNavigli:2014a}, taxonomy learning \cite{espinosa2016extasem}, knowledge-base completion \cite{bordes2013translating}, information extraction \cite{dellibovi-espinosaanke-navigli:2015:EMNLP},  or sense-based semantic similarity \cite{BudanitskyHirst:06,Pilehvaretal:2013,iacobacci:2015}, to name a few.

\subsection{Open problems and future work}

This last session provides a summary of possible directions of future work on semantic sense representation. We discuss various problems associated with the current representation approaches and propose lines of research in order to effectively apply sense representations in natural language understanding tasks.

\section{Instructors}

\paragraph{Jos\'e Camacho Collados} is a Google Doctoral Fellow and PhD student at the Sapienza University of Rome\footnote{\url{http://wwwusers.di.uniroma1.it/~collados/}}, working under the supervision of Prof. Roberto Navigli. His research focuses on Natural Language Processing and on the area of lexical semantics in particular. He has developed \textsc{Nasari}\footnote{\url{http://lcl.uniroma1.it/nasari/}} \cite{camachocollados:2016aij}, a novel semantic vector representation for concepts based on Wikipedia that features a high coverage of named entities and has been successfully used on different NLP tasks. Jos\'e has also worked on the development of evaluation benchmarks for word and sense representations \cite{camacho2015framework,camachocolladosrepeval:2016} and is the co-organizer of a SemEval 2017 task on multilingual and cross-lingual semantic similarity. His background education includes an Erasmus Mundus Master in Natural Language Processing and Human Language Technology and a 5-year BSc degree in Mathematics.

\paragraph{Ignacio Iacobacci} is a PhD student at the Sapienza University of Rome\footnote{\url{https://iiacobac.wordpress.com/}}, working under the supervision of Prof. Roberto Navigli. His research interests lie in the fields of Machine Learning, Natural Language Processing, Neural Networks. He is currently working on Word Sense Disambiguation and Distributional Semantics. Ignacio presented SensEmbed\footnote{\url{http://lcl.uniroma1.it/sensembed/}} at ACL 2015 \cite{iacobacci:2015}, a novel approach for word and relational similarity built from exploiting semantic knowledge for modeling arbitrary word senses in a large sense inventory. His background includes a MSc. in Computer Science and 8 years as a developer including 4 years as a Machine Learning - NLP specialist.

\paragraph{Roberto Navigli} is an Associate Professor in the Department of Computer Science at La Sapienza University of Rome and a member of the Linguistic Computing Laboratory\footnote{\url{http://wwwusers.di.uniroma1.it/~navigli/}}. His research interests lie in the field of Natural Language Processing, including: Word Sense Disambiguation and Induction, Ontology Learning, Knowledge Representation and Acquisition, and multilinguality. In 2007 he received a Ph.D. in Computer Science from La Sapienza and he was awarded the Marco Cadoli 2007 AI*IA national prize for the Best Ph.D. Thesis in Artificial Intelligence. In 2013 he received the Marco Somalvico AI*IA prize, awarded every two years to the best young Italian researcher in Artificial Intelligence. He is the creator and founder of BabelNet\footnote{\url{http://www.babelnet.org}} \cite{NavigliPonzetto:12aij}, both a multilingual encyclopedic dictionary and a semantic network, and its related project Babelfy\footnote{\url{http://www.babelfy.org}} \cite{Moroetal:14tacl}, a state-of-the-art multilingual disambiguation and entity linking system. He is also the Principal Investigator of MultiJEDI\footnote{\url{http://multijedi.org/}}, a 1.3M euro 5-year Starting Grant funded by the European Research Council and the responsible person of the Sapienza unit in LIDER, an EU project on content analytics and language technologies. Moreover, he is the Co-PI of "Language Understanding cum Knowledge Yield" (LUcKY), a Google Focused Research Award on Natural Language Understanding.

\paragraph{Mohammad Taher Pilehvar} is a Research Associate in the Language Technology Lab of the University of Cambridge\footnote{\url{http://www.pilevar.com/taher/}} where he is currently working on NLP in the biomedical domain. Taher completed his PhD in 2015 under the supervision of Prof. Roberto Navigli. Taher's research lies in lexical semantics, mainly focusing on semantic representation, semantic similarity, and Word Sense Disambiguation. He has co-organized three SemEval tasks \cite{jurgens2014semeval,jurgens-pilehvar:2016:SemEval} and has authored multiple conference and journal papers on semantic representation and similarity in top tier venues. He is the first author of a paper on semantic similarity that was nominated for the best paper award at ACL 2013 \cite{PilehvarNavigli:2013}.

\section*{Acknowledgments}
The authors gratefully acknowledge the support of the ERC Starting
  Grant MultiJEDI No.\ 259234.

\bibliography{acl2016}
\bibliographystyle{acl2016}

\end{document}